\documentclass{article}

\usepackage{arxiv}

\usepackage[utf8]{inputenc} 
\usepackage[T1]{fontenc}    
\usepackage{hyperref}       
\usepackage{url}            
\usepackage{booktabs}       
\usepackage{amsfonts}       
\usepackage{nicefrac}       
\usepackage{microtype}      
\usepackage{lipsum}         
\usepackage{fancyvrb}       
\usepackage{comment}        
\usepackage{wrapfig}        
\usepackage{cleveref}       
\usepackage[nottoc]{tocbibind}  
\usepackage[export]{adjustbox}  

\usepackage{listings}
\usepackage{multirow}

\usepackage{tocloft}
\usepackage{microtype}
\usepackage{algorithmic}
\usepackage{algorithm}

\usepackage{todonotes}

\newcommand{\version}{1.0}

\title{Design and Implementation of TAG:\\ A Tabletop Games Framework}

\author{Raluca D. Gaina\textsuperscript{*}, Martin Balla\textsuperscript{*}, Alexander Dockhorn\textsuperscript{*}, \\
\textbf{Ra\'{u}l Montoliu\textsuperscript{†}, Diego Perez-Liebana\textsuperscript{*}}\\
\textsuperscript{*}School of Electronic Engineering and Computer Science,  Queen Mary University of London, UK\\
\textsuperscript{†}Insitute of New Imaging Technologies, Jaume I University, Castellon, Spain}

\begin{document}
\maketitle

\begin{abstract}

This document describes the design and implementation of the Tabletop Games framework (TAG), a Java-based benchmark for developing modern board games for AI research. TAG provides a common skeleton for implementing tabletop games based on a common API for AI agents, a set of components and classes to easily add new games and an import module for defining data in JSON format. At present, this platform includes the implementation of seven different tabletop games that can also be used as an example for further developments. Additionally, TAG also incorporates logging functionality that allows the user to perform a detailed analysis of the game, in terms of action space, branching factor, hidden information, and other measures of interest for Game AI research. The objective of this document is to serve as  a central point where the framework can be described at length. TAG can be downloaded at:

\centering\url{https://github.com/GAIGResearch/TabletopGames}
\end{abstract}

\keywords{Tabletop Games \and General Game AI \and TAG \and Statistical Forward Planning}
\rule{\textwidth}{2pt}

\begin{center}
\includegraphics[width=0.48\textwidth, fbox]{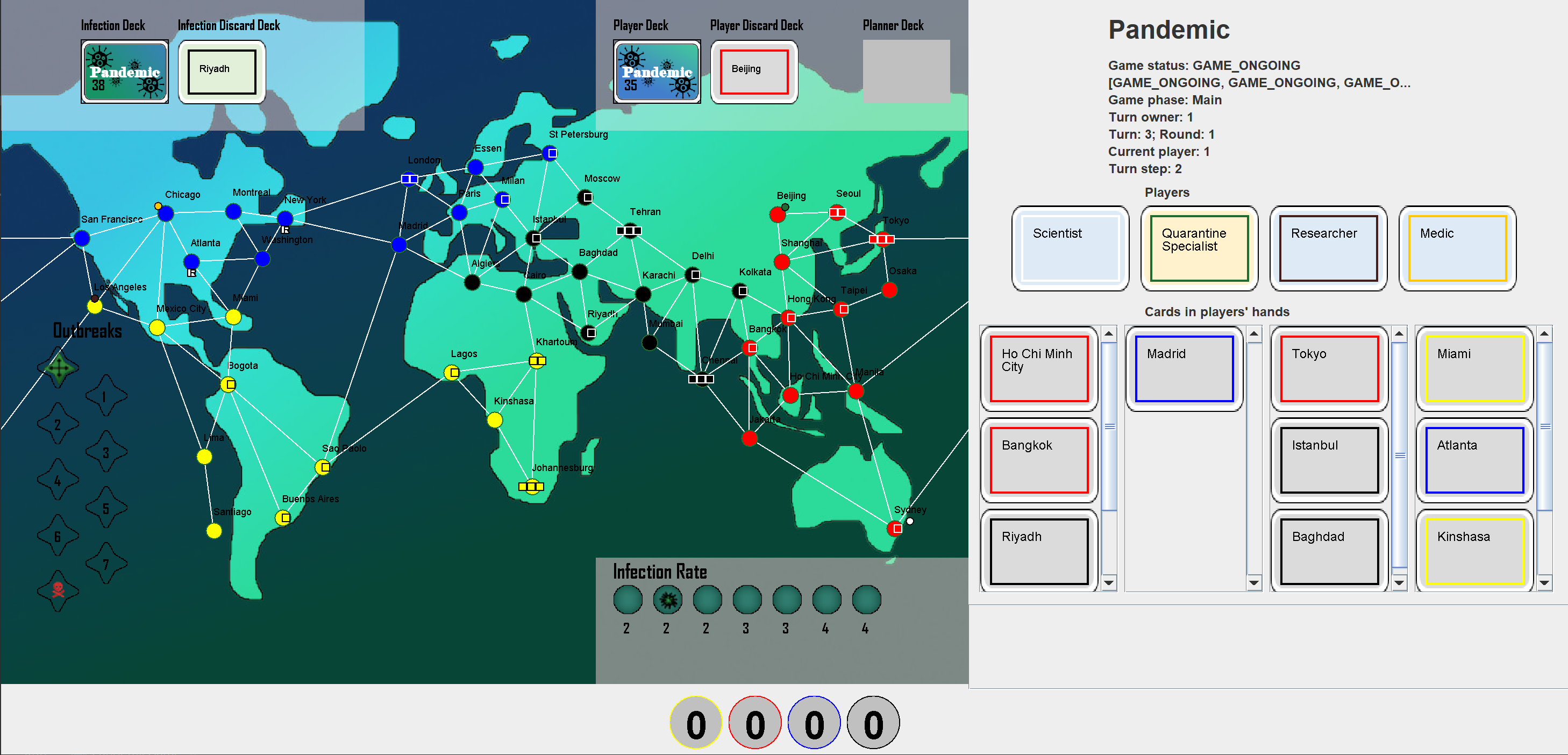}
\includegraphics[width=0.48\textwidth, fbox]{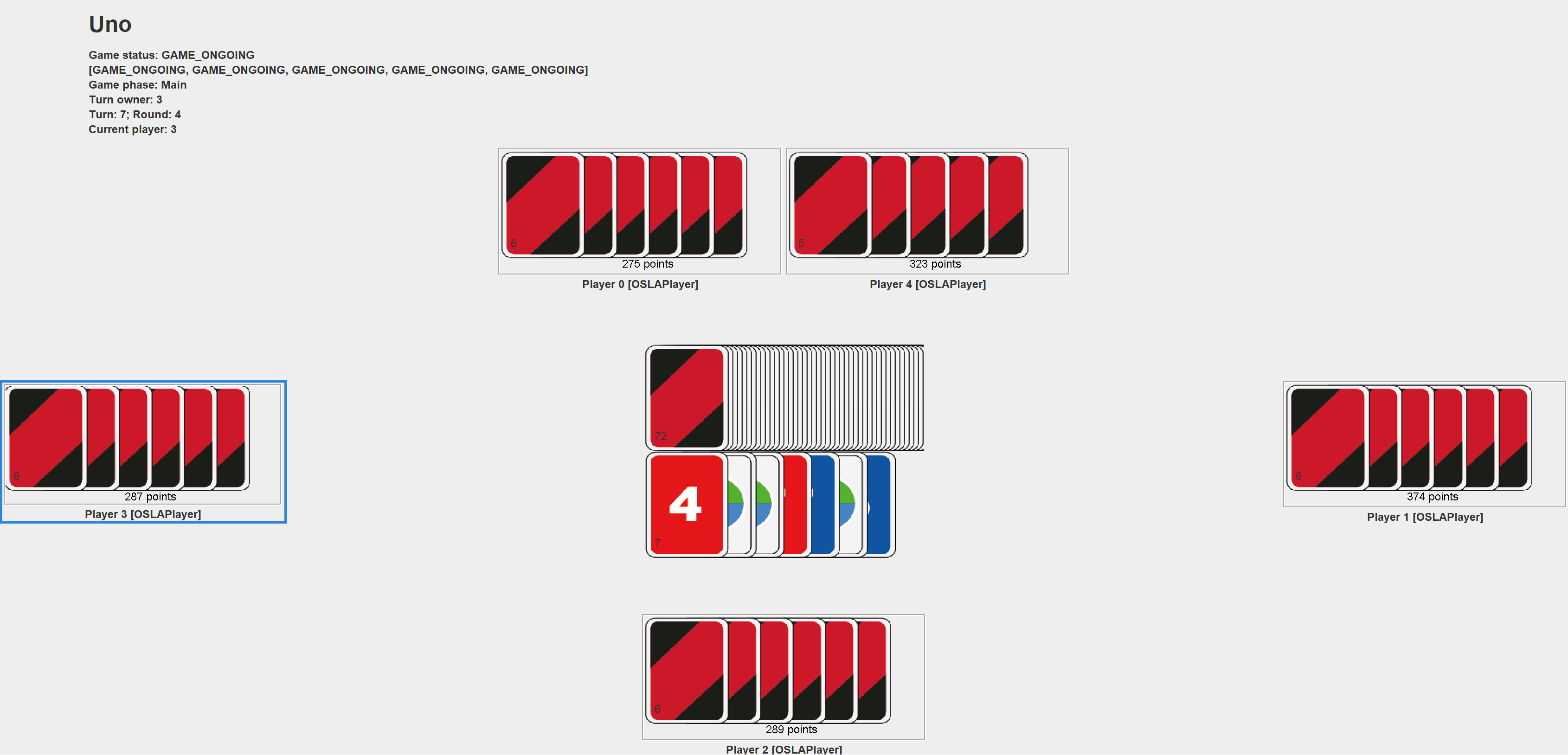}
\includegraphics[width=0.48\textwidth, fbox]{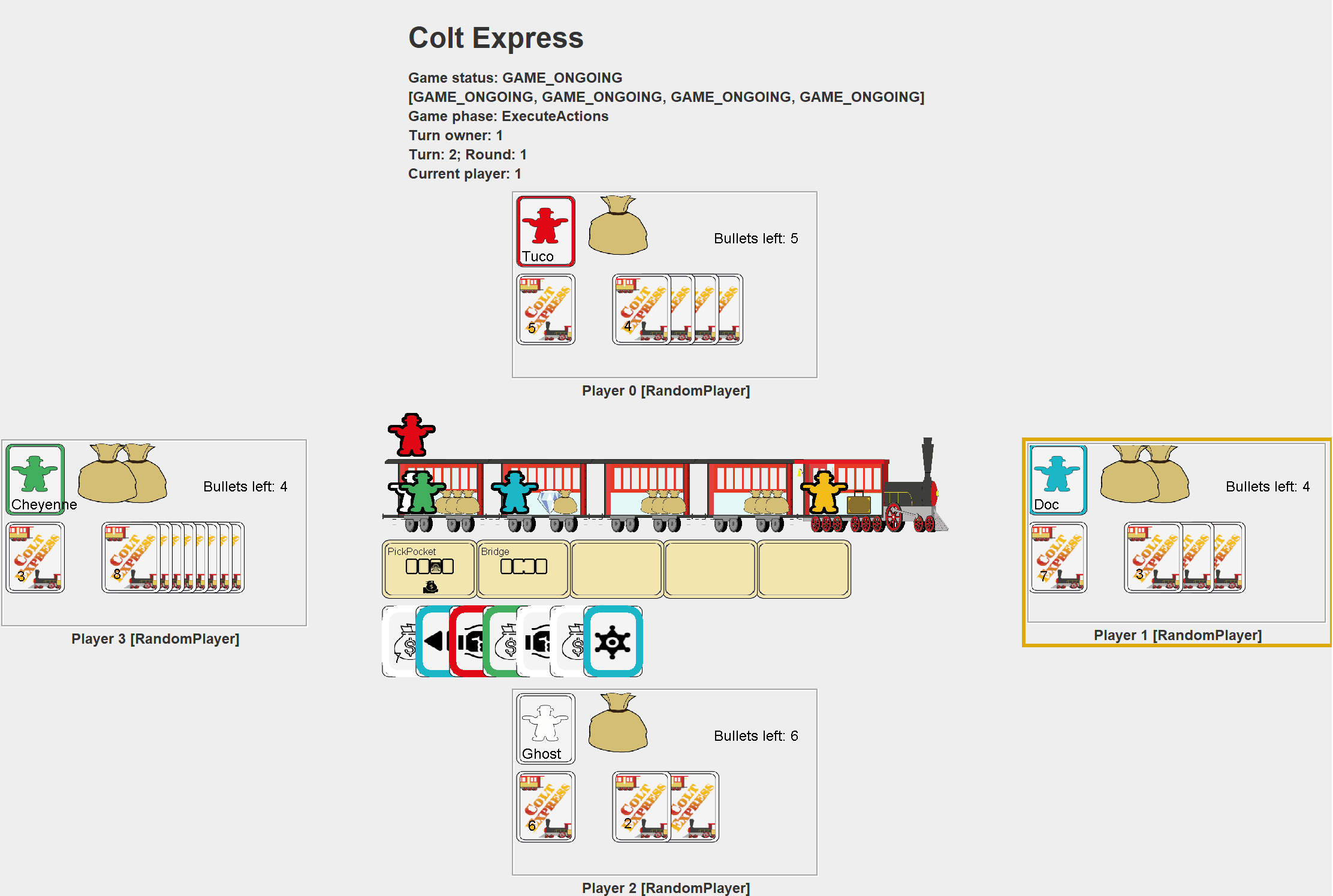}
\includegraphics[width=0.48\textwidth, fbox]{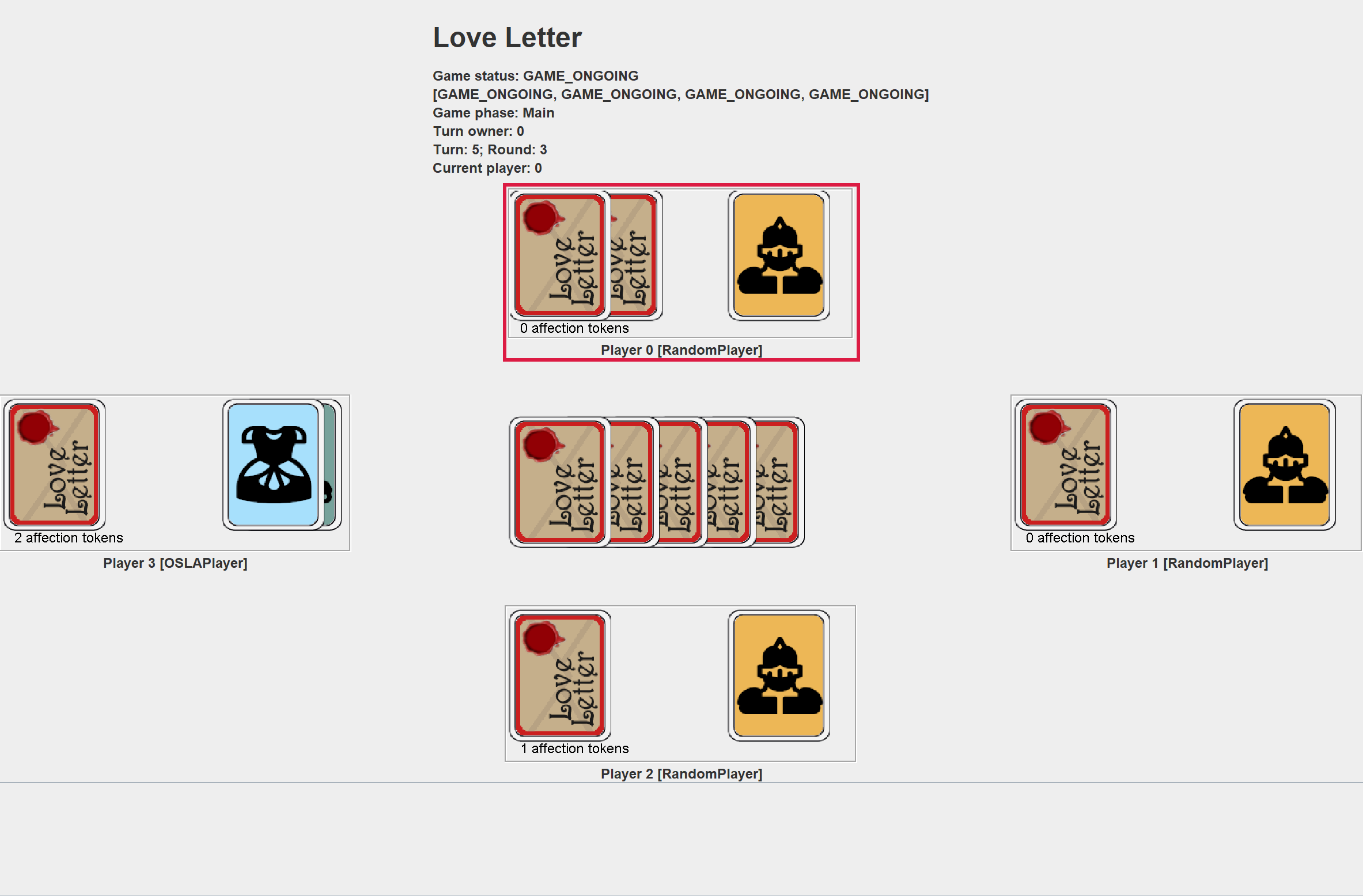}
\end{center}

\newpage

\rule{\textwidth}{2pt}
{\tableofcontents}
\rule{\textwidth}{2pt}

\newpage

\section{How to use this Document?}

This paper serves the purpose of a living documentation for TAG - a Tabletop Games framework for studying computational intelligence in this type of games. This document will be continuously updated with the new developments of the framework. When citing this paper please make sure to specify its version since the paper will be updated over time.

\section{TAG and Tabletop Games}

TAG was designed to capture most of the complexity that modern tabletop games provide, with a few games implemented already and more in progress. Our framework includes handy definitions for various concepts and components common across many tabletop games~\cite{engelstein2019building}.

We define an \textbf{action} as an independent unit of game logic that modifies a given game state towards a specific effect (e.g. player draws a card; player moves their pawn). These actions are executed by the game players and are subject to certain \textbf{rules}: units of game logic, part of a hierarchical structure (a game flow graph). Rules dictate how a given game state is modified and control the flow through the game graph (for instance, checking the end of game conditions and the turn order). This \textbf{turn order} defines which player is due to play at each time, possibly handling player reactions forced by actions or rules. At a higher level, games can be structured in \textbf{phases}, which are time frames where specific rules apply and/or different actions are available for the players.

All tabletop games use \textbf{components} (game objects sharing certain properties), whose state is modified by actions and rules during the game. TAG includes several predefined components to ease the development of new games, such as tokens (a game piece of a particular type), dice (with N sides), cards (with text, images or numbers), counters (with a numerical value), grid and graph boards. Components can also be grouped into collections: an \textbf{area} groups components in a map structure in order to provide access to them using their unique IDs, while a \textbf{deck} is an ordered collection (list) with specific interactions available (e.g. shuffle, draw, etc.). Both areas and decks are considered components themselves. A summary of the components follows:

\begin{itemize}
    \item \textbf{Token}: A game piece of a particular type, usually with a physical position associated with it.
    \item \textbf{Die}: A die has a number of sides N associated with it and can be rolled to obtain a value between 1 and N (inclusive).
    \item \textbf{Card}: A card usually has text, images or numbers associated with any of its 2 sides, and is the most common type of component used in decks.
    \item \textbf{Counter}: An abstract concept used to keep track of a particular variable numerical value; usually represented on a board with tokens used to mark the current value, but recognized as a separate object in this framework. It has a minimum, maximum and current value associated with it, where the current value can vary between the minimum (inclusive) and maximum (inclusive).
    \item \textbf{Graph board}: A graph representation for a board, as a collection of several board nodes connected between each other.
    \item \textbf{Board node}: A node in a graph board which keeps track of its neighbours (or connections) in the board.
    \item \textbf{Grid board}: A 2D grid representation of a board, with a width and height associated with it. It can hold elements of any type.
\end{itemize}

The structure of TAG consists of several packages:
\begin{itemize}
    \item \texttt{core}: All core framework functionality, including all abstract classes to be extended by game implementations.
    \item \texttt{evaluation}: Classes for running tournaments and evaluations of games or AI players.
    \item \texttt{games}: Specific implementations of abstract classes for each game, each grouped in its own package.
    \item \texttt{gui}: Generic Graphical User Interface (GUI) helper classes, including the \texttt{PrototypeGUI} class which can be used with any in-progress game. Each game can extend this to implement customized functionality in their GUIs, for better player interaction and thus a better player experience.
    \item \texttt{players}: All players (human and AI) available.
    \item \texttt{utilities}: Various utility classes and generic functionality shortcuts.
\end{itemize}

Additionally, the folder \texttt{data} (outside of the source files) stores assets for the games, each in their own folders. Most of these refer only to images used in the interfaces, but may include JSON files and other data as well (e.g. \textit{Pandemic} stores board and card information here as well).

The TAG framework brings together all of the concepts and components described previously and allows quick implementation and prototyping of new games. To this end, a flexible API is provided for all functionality needed to define a game, with multiple abstract classes that can be extended for specific implementations. The framework already provides some generic functionality: ready-made components, rules, actions, turn orders and game phases. These are all represented through several game objects found in the \texttt{core} package, which can be instantiated in game implementations for immediate use. Additionally, TAG includes a fully functional game loop and a prototyping GUI. The GUI allows users to start interacting with the games as soon as they have the two main classes required set up: a \textit{Game State} (\textbf{GS}) class and a \textit{Forward Model} (\textbf{FM}) class. 

\textbf{GS} is a container class, including all variables and game components which would allow one to describe one specific moment in time. It defines access methods in the game state to retrieve existing game components, make custom and partially observable copies of the state, and define an evaluation function that can be used by the playing agents. We strongly recommend that game components (i.e. extending the \textit{Component} class) are used to describe a game state as much as possible (e.g. using Counters instead of integer variables to keep track of numbers in the game), in order for general AI players to have an easy common access to understand game states. 

\textbf{FM} encompasses the logic of the game: performs the game setup (by setting the GS components and variables to their initial state and/or reading from files and creating components); defines what actions players can take in a particular game state; applies the effect of player actions (received one at a time; if a simultaneous-action game is to be implemented, we recommend waiting to receive all before modifying the game state) and any other game rules applicable; uses a turn order to decide which player is due to play next, and checks for any end of game conditions (and setting the game status and player results in the game state appropriately if the game is over). The FM is available to AI players for game simulations.

For each game, users can further implement specific actions, rules, turn orders, game parameters (for easy modification of game mechanics), a GUI and provision of game data. The last is useful when the game requires large amounts of data such as tile patterns, cards and board node connections, and it is provided via JSON files. A full guide on using the framework and implementing new games is available in the wiki provided with the code\footnote{https://github.com/GAIGResearch/TabletopGames/wiki}.

TAG's game-loop is presented in \Cref{alg:game-loop}. Given a list of agents and parameters of the game to be played, the framework performs an initial setup of the game state ($s_0$) and the game's forward model (FM). While the game state is not terminal, the turn order selects the next player to act. To ensure partial observability, we generate an observation object for the current agent which hides the state of unobserved components. Hence, a list of available actions is generated and the agent is queried to provide the next action to be executed ($a_t$). Finally, the forward model is used to modify the game state given the action (producing $s_{t+1}$), and the graphical user interface is updated accordingly.

\begin{algorithm}[t] 
\caption{Overview Game Loop} 
\label{alg:game-loop} 
\begin{algorithmic} 
    \REQUIRE list of agents, game parameters $gp$
    \ENSURE win rate statistics
    \vspace{1em}
    \STATE $s_0$, FM = \textsc{setupGame}($gp$)
    \WHILE{\textbf{not} \textsc{isTerminal}($s_t$)}
        \STATE $agent$ $\gets$ \textsc{getCurrentPlayer}($s_t$)
        \STATE observation $\gets$ \textsc{getObservation}($s_t$, $agent$)
        \STATE actions $\gets$ FM.\textsc{getAvailableActions}($s_t$, $agent$)
        \STATE $a_t$ $\gets$ $agent$.\textsc{getAction}(observation, actions)
        \STATE $s_{t+1} \gets$ FM.\textsc{next}($s_t$, $a_t$)
        \STATE GUI.\textsc{update}()
    \ENDWHILE
\end{algorithmic}
\end{algorithm}

\section{Implementing Games in TAG}

This section dives deeper into the structure of the framework. TAG can be used to implement new games, for which it provides a set of abstract classes and interfaces to ease the development. Although there is no hard requirement to implement games in a particular way, the recommended set of classes to implement is described here.

In order to implement a new game in TAG, the following core modules can be implemented:

\subsection{Game State} \label{sec:imp:gs}

This core class includes all the information about the components that form the state of the game. New game states can extend the class \texttt{core.AbstractGameState}, implementing the following methods:

\begin{itemize}
    \item \textit{\_getAllComponents()}: returns a list of all the components of the game state. The method is called after game setup, so you may assume all components have already been created. Decks and Areas have all of their nested components automatically added.
    \item \textit{\_copy(int playerId)}: defines a reduced, player-specific, copy of your game state. This includes only those components (or parts of the components) which the player with the given ID can currently see. For example, some decks may be face down and unobservable to the player. All of the components in the observation should be \textit{copies} of those in the game state, as the player is not prohibited from modifying the state it receives (copies ensures the real objects in the game are not affected in any way).
    \item \textit{\_reset()}: resets any variables of the game state to their state before FM initialisation.
    \item \textit{\_getScore(int playerId)}: implements a heuristic function that returns a numerical value for the current game state, given a specific player - the bigger this value, the better the state.
\end{itemize}

Additionally, game state classes can implement the \texttt{core.interfaces.IFeatureRepresentation} interface, which provides an interface for methods that can be used to extract further generic information about the game state, in terms of abstract features. Game state classes can also implement the \texttt{core.interfaces.IVectorObservation} interface, which allows AI players to receive a vector observation from the game state. 

Appendix~\ref{app:loveletter:gs} shows an example of a Game State implementation for the game Love Letter.

\subsection{Forward Model}

The forward model class is in charge of setting up the initial game state and advancing the game state when provided with an action to be executed by one of the players. In order to implement a forward model, extend the class \texttt{core.AbstractForwardModel} and implement the following methods:

\begin{itemize}
\item \textit{\_setup(AbstractGameState firstState)}: perform the initial game setup according to the game rules, initialising all components and variables in the given game state (e.g. give each player their starting hand, place tokens on the board etc.). 
\item \textit{\_next(AbstractGameState currentState, AbstractAction action)}: apply the given action to the given game state, execute any other non-action-dependent game rules (e.g. related to game phase changes), check for game end conditions (setting the game status and player results in the given state) and move to the next player if required.
\item \textit{\_computeAvailableActions(AbstractGameState gameState)}: return a list with all actions available for the current player, in the context of the game state object.
\item \textit{\_copy()}: return a new instance of the Forward Model object with any necessary variables copied. Use a new random seed (e.g. \texttt{System.currentTimeMillis()} for initialisation of the copy object).
\item \textit{endGame()}: override this function if the game requires any extra end of game computation (e.g. to print end of game results).
\end{itemize}

Appendix~\ref{app:loveletter:fm} shows an example of a Forward Model implementation for the game Love Letter.

\textbf{Note:} Forward model classes can instead extend from the abstract class \texttt{core.rules.AbstractRuleBasedForwardModel}, if they wish to use the rule-based system instead; this class provides basic functionality and documentation for using rules, and an already implemented \texttt{\_next()} function.  

\subsection{Actions}

TAG provides a series of simple actions that are common for many games, such as shuffling decks (or, more generally, shuffling groups of components), drawing components from one deck to another, replacing components, re-arranging decks or placing tokens in grid boards. The games currently included in the framework use these actions to manipulate game components such as decks and counters in the game. 

It is also possible to define game specific actions for new games. For this, a new action class needs to be created that extends from \texttt{core.actions.AbstractAction}. Two main methods can be overriden:

\begin{itemize}
\item \textit{execute(AbstractGameState gs)}: this method executes the action the class represents, applying its effect to the given game state. Within this method, any component that has been registered with a unique ID can be retrieved from the game state with the function \texttt{AbstractGameState.getComponentById(int id)}.
\item \textit{copy()}: this method returns an exact copy of this action.
\end{itemize}

\textbf{Important:} Actions should not hold any reference to other objects to avoid cross-referencing in the game and prevent issues with deep copies. Unique component IDs should be passed instead into action classes constructors. 

Examples of game-related actions for Love Letter can be seen in Appendix~\ref{app:loveletter:actions}

\subsection{Other modules}

Additionally, some other classes should also be implemented or added to, following the framework's structure:

\begin{itemize}
    \item \textbf{Game Type:} TAG allows to categorize its games into different internal collections of games of the similar type. This allows the user, for instance, to retrieve games with certain mechanics or number of players within the framework. The class \texttt{games.GameType} includes several methods that can be added to so that a new game is known and search-able within the framework. In order to incorporate a game to the framework in this way, you should add the following:
    \begin{itemize}
        \item Add a new \texttt{enum} value for your game, defining the minimum and maximum number of players, a list of categories the game belongs to, and a list of mechanics the game uses (both categories and mechanics can be already existing ones or newly defined options).
        \item In the method \texttt{stringToGameType(String game)}, add a new case for your game so that the string name can be converted to the correct game type.
        \item In the method \texttt{createGameInstance(int nPlayers, long seed)}, add a new case for your game, creating the appropriate instances of the game state and forward model needed to run, given the number of players taking part in the game and the random seed.
        \item In the method \texttt{createGUI(AbstractGameState gameState, ActionController ac)}, add a new case for your game (if GUI is implemented, the PrototypeGUI may be used otherwise), creating the appropriate instance of GUI, given an initial game state and the action controller managing user interactions within the GUI.
    \end{itemize}
    \textbf{Important:} doing this step is necessary for using the main methods of running games in the framework in the \texttt{core.Game} class and game analysis. 
    \item \textbf{GUI:} The abstract class \texttt{core.AbstractGUI} can be extended to indicate not only how the game should be displayed in a graphical user interface, but also to specify what bits should be occluded to respect partial observability of the game state for human players. The class \texttt{gui.PrototypeGUI} and the other GUI helper classes in package \texttt{gui} can be used as a simple template to be customized depending on necessary visualisations and interactions.
    \item \textbf{Game Parameters:} These allow tweaking the game behaviour and rules when these depend on numerical or discrete values. For instance, the number of cards in a hand of Uno or the number of outbreaks allowed before losing a game of Pandemic can be specified this way. This provides a centralized point for the parameters of a game that define its behaviour and allow play-testing different versions of the same game easily. In order to implement game parameters, a parameters class for a specific game must extend \texttt{core.GameParameters} or implement \texttt{core.interfaces.ITunableParameters} (which allows game parameters to be randomized or tuned).
    \item \textbf{Turn Order:} The order in which players make moves in the game can be specified by extending the abstract class \texttt{core.turnorders.TurnOrder}, which provides functionality to establish the starting and current player and advance the turn to the next player to move. Default turn orders such as alternate (one after another until all players have played) and reactive (which adds the possibility of a player to act out of turn) are included in the framework and can be reused for new games.
    \item \textbf{Constants:} For efficiency reasons, you may declare constants of a given game for quick access through the framework. The class \texttt{pandemic.PandemicConstants} shows an example of how to specify game-related constants for a game, and how to use the framework's Hash utility to provide direct access to certain game components.
\end{itemize}

The following classes can further be implemented, but they are not required for essential game functionality, and may not be needed at all depending on the game implemented:

\begin{itemize}
    \item \textbf{Game (Optional):} Extending \texttt{core.Game}, allows to specify the types of parameters, game state, GUI and forward model for a specific game. This class can also be used to add the entry point to run the game with some AI agents or humans.
    \item \textbf{Game Data (Optional):} Some tabletop games need extra data information, which may come in the form of specific card sections (costs, effects, categories, etc.), tile patterns, etc. This data can be specified in json format files for each component type and read with the json parser incorporated in the framework. Alternatively, it is possible to add own format and parser code. The abstract class \texttt{core.AbstractGameData} can be extended to incorporate data reading for the particular game.
    \item \textbf{Game Phase (Optional):} Some games can be structured in phases, we different rules and available actions apply. The \texttt{core.AbstractGameState} class includes a definition for some basic default game phases (main, player reaction, end), which could be expanded in order to provide specific phases for new games.
\end{itemize}

\section{Games}\label{sec:games}

There are currently 7 games implemented in the framework, varying from very simple test games (\textit{Tic-Tac-Toe}) to strategy games (\textit{Pandemic}~\cite{game:pandemic}), as well as diverse challenges for AI players. A few games are currently in active development (\textit{Descent}~\cite{game:descent}, \textit{Carcassonne}~\cite{game:carcassonne} and \textit{Settlers of Catan}~\cite{game:catan}), and many more are in the project's backlog, including games from other frameworks to allow for easy comparison. Further development plans also include adding easy to use functionality for wrapping external games, so that a direct comparison can be carried out under the same conditions without the need to re-implement games under our framework's full restrictions.

All games implemented can be found in the \texttt{games} package, each registered in the \texttt{games.GameType} class; this class allows specifying properties for each game, to allow for automatic listing for experiments (e.g. a list of all games with the ``cooperative'' tag). We highlight next some particularities of the games currently implemented in the framework. 

\subsection{Tic-Tac-Toe}


$2$ players alternate placing their symbol in a $N\times N$ grid until one player completes a line, column or diagonal and wins the game; if all cells in the grid get filled up without a winner, the game is a draw. This is the simplest game included in the framework, meant to be used as a quick reference for the minimum requirements to get a game up and running.  Its implementation makes use of mostly default concepts and components, but it implements a scoring heuristic and a custom GUI for an easier interaction given the specific game mechanics. 

\subsection{Love Letter~\cite{game:loveletter}}

\begin{figure*}[!ht]
    \centering
    \includegraphics[width=0.8\textwidth]{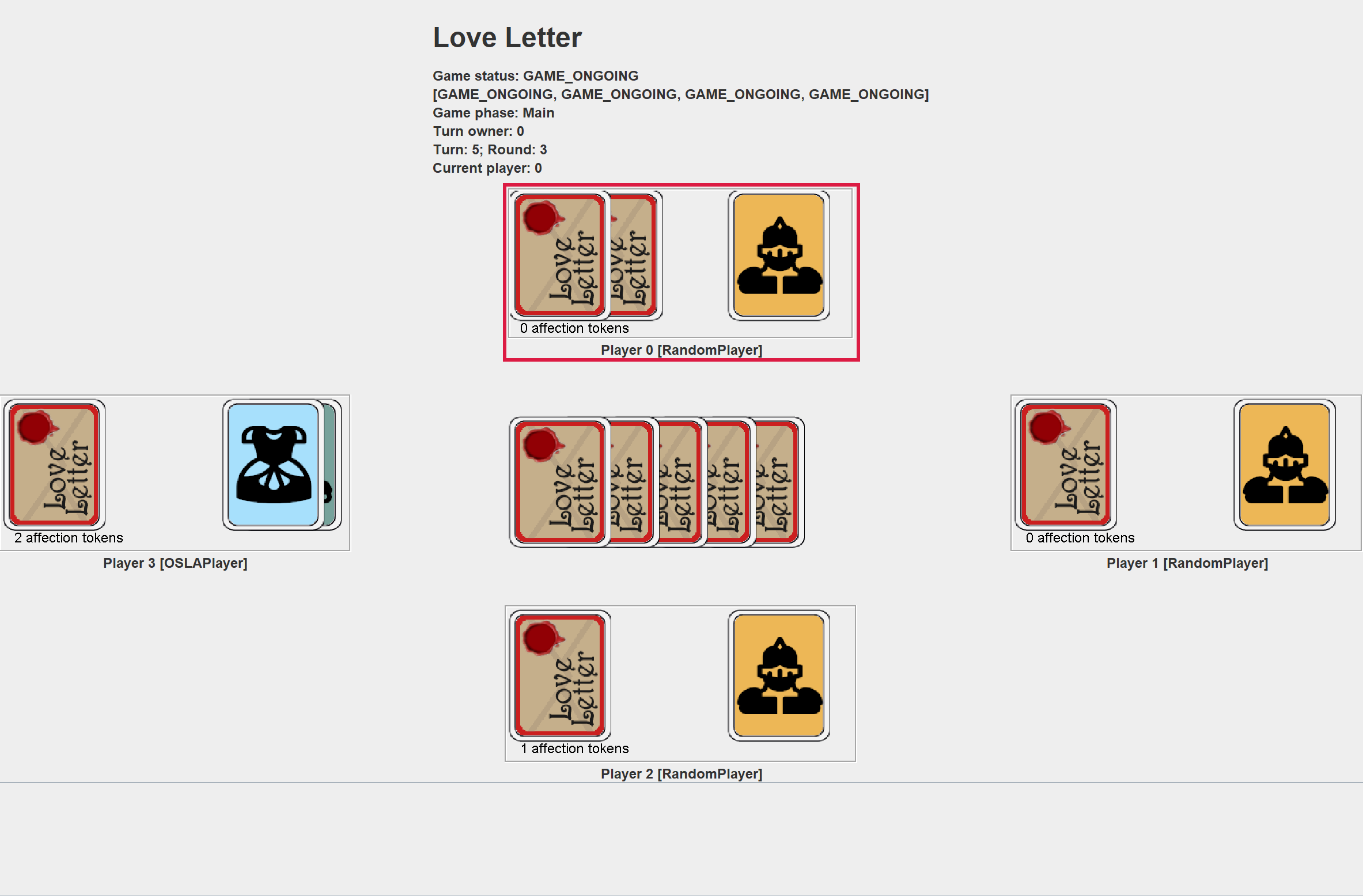}
    \caption{Screenshot of Love Letter in TAG.}
    \label{fig:ll}
\end{figure*}

$2$ to $4$ players start the game with one card each, representing a character, a value and a unique effect. A second card is drawn at the start of each turn, one of which must be played afterwards. 
After the last card of the deck is drawn, the player with the highest valued card wins the current round. A player wins the game after winning 5 rounds. \textit{Love Letter} features partial observability, asymmetric and changing player roles and a point system over several rounds. Figure~\ref{fig:ll} shows an example game state.

\subsection{Colt Express~\cite{game:coltexpress}}

\begin{figure*}[!ht]
    \centering
    \includegraphics[width=0.8\textwidth]{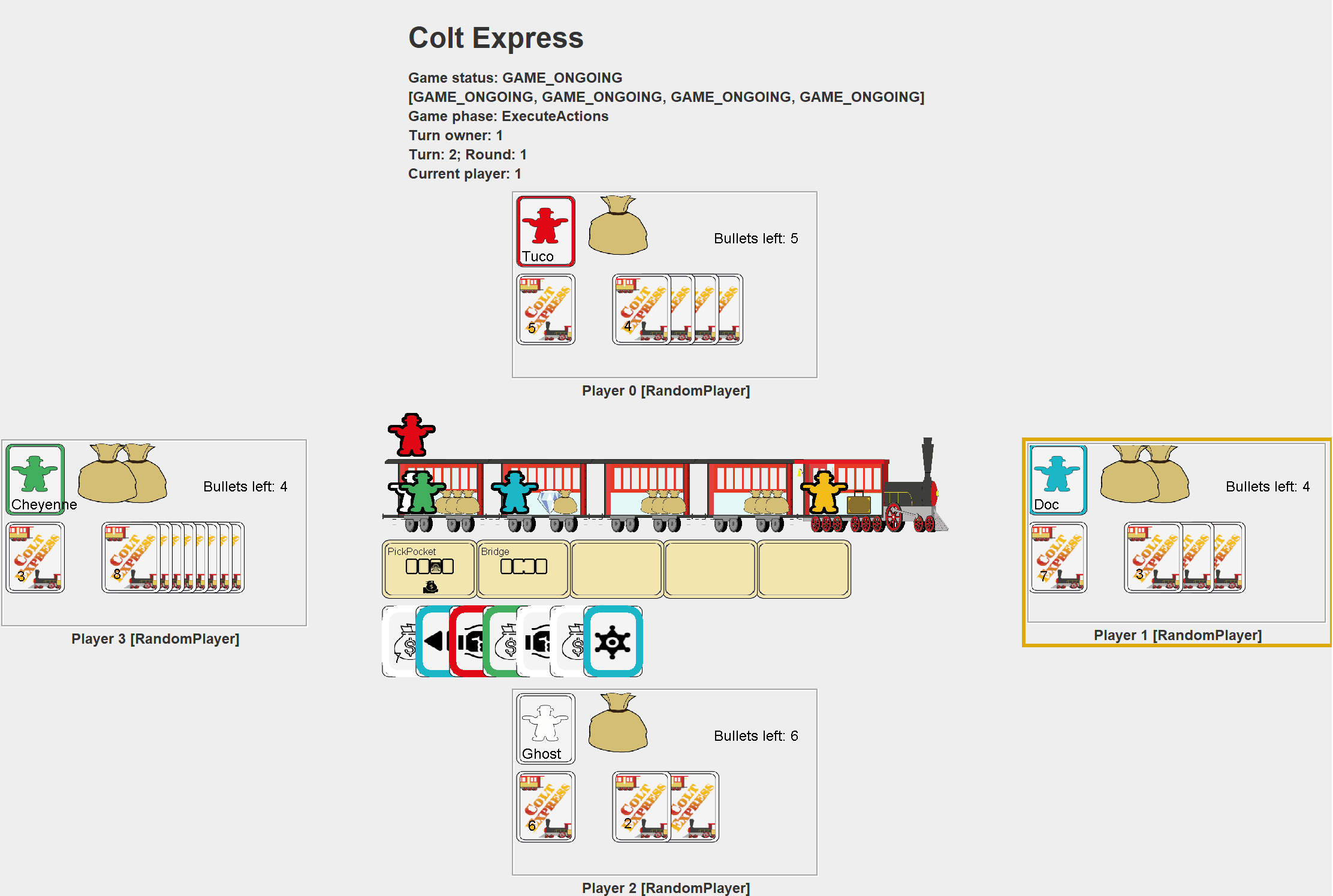}
    \caption{Screenshot of Colt Express in TAG.}
    \label{fig:ce}
\end{figure*}

$2$ to $6$ players control a bandit each, with a unique special ability. Their goal is to collect the most money while traversing the two-level compartments in a train and avoiding the sheriff (a non-player character moved by players and round card events). The game consists of several rounds, each with a planning (players play action cards) and an execution (cards are executed in the same order) phase.
During the former, players play action cards, which are executed in the same order in the latter phase. 
This processing scheme forces players to adapt their strategy according to all the moves already played, in an interesting case of partial observability and non-determinism: the opponents' type of action may be known (sometimes completely hidden in a round), but not how it will be executed. Additionally, the overall strategy should be adapted to a bandit's unique abilities. Those actions can move the player's character along the compartments, interact with neighbouring players, or collect money. The game ends after all rounds are played, the player with the most money being the winner. Figure~\ref{fig:ce} shows an example game state.

\subsection{Uno~\cite{game:uno}}

\begin{figure*}[!ht]
    \centering
    \includegraphics[width=0.8\textwidth]{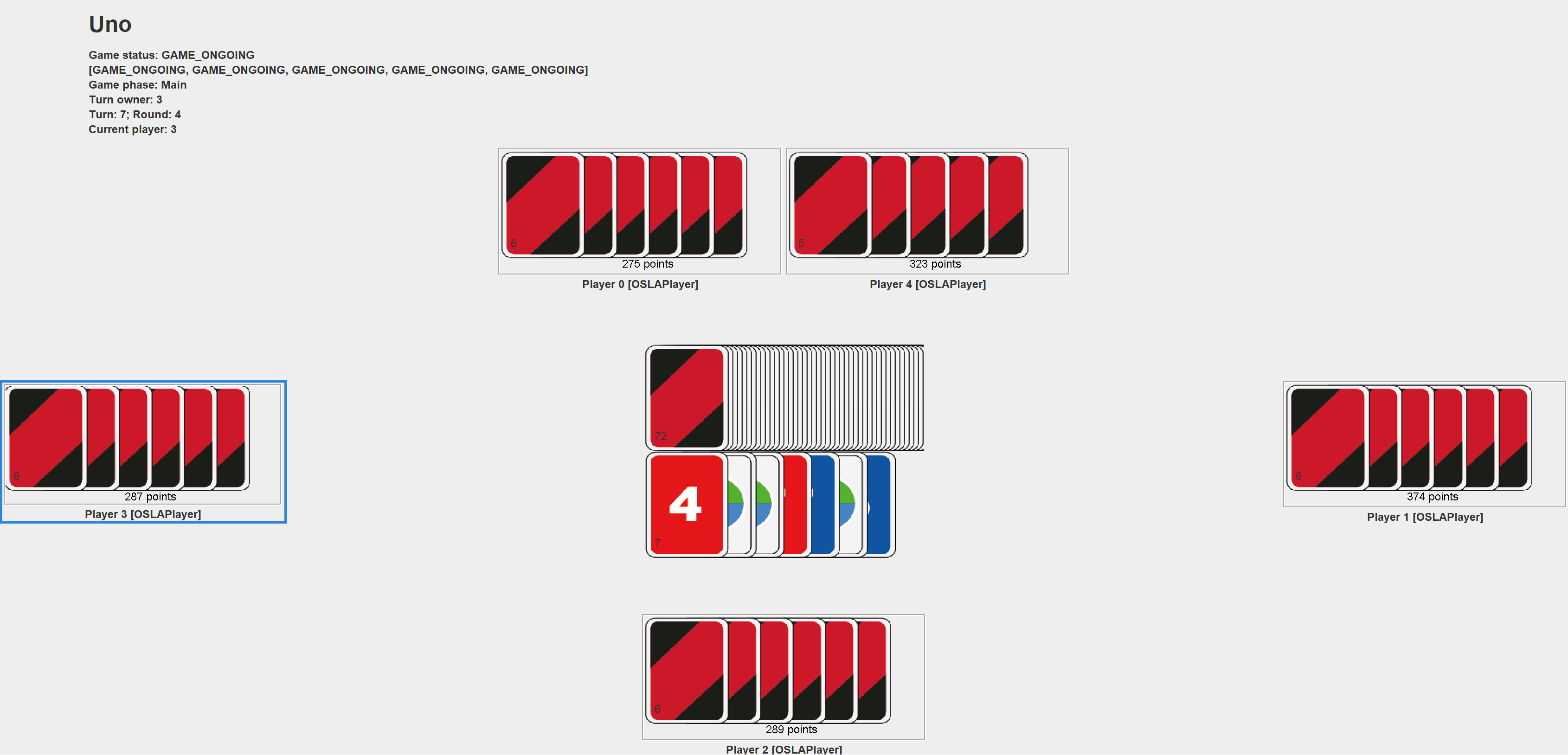}
    \caption{Screenshot of Uno in TAG.}
    \label{fig:uno}
\end{figure*}

The game consists of coloured cards with actions or numbers. Numbered cards can only be played in case either the colour or the number matches the newest card on the discard pile. Action cards let $2$ to $10$ players draw additional cards, choose the next colour to be played or reverse the turn order. A player wins after gaining a number of points over several rounds (computed as the sum of all other players' card values). \textit{Uno} features stochasticity, partial observability and a dynamically changing turn order.
This game has the potential of being the longest game in the framework since players need to draw new cards in case they cannot play any. Figure~\ref{fig:uno} shows an example game state.

\subsection{Virus!~\cite{game:virus}}

$2$ to $6$ players have a body each that consists of four organs, which can be: infected (by an opponent playing a virus card), vaccinated (by a medicine card), immunised (by 2 medicine cards) or destroyed (by opponents playing 2 consecutive virus cards). The winner is the first player who forms a healthy and complete body. \textit{Virus!} features stochasticity and partial observability, with the draw pile and opponents' cards being hidden.

\subsection{Exploding Kittens~\cite{game:kittens}}

\begin{figure*}[!ht]
    \centering
    \includegraphics[width=0.8\textwidth]{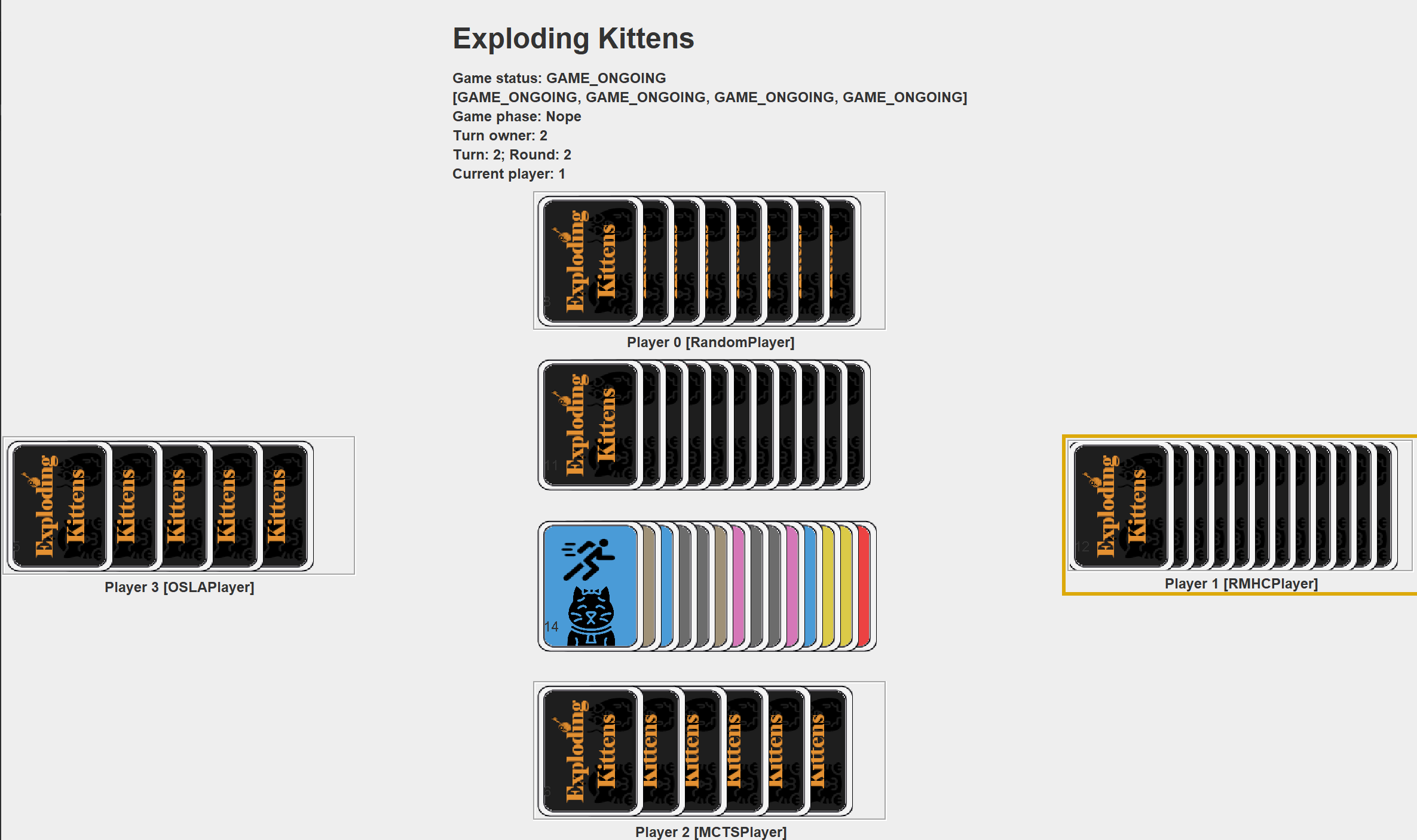}
    \caption{Screenshot of Exploding Kittens in TAG.}
    \label{fig:ek}
\end{figure*}

$2$ to $5$ players try to avoid drawing an exploding kitten card while collecting other useful cards. Each card gives a player access to unique actions to modify the game-state, e.g. selecting the player taking a turn next and shuffling the deck.
This game features stochasticity, partial observability and a dynamic turn order with out-of-turn actions: in contrast to previous games, \textit{Exploding Kittens} keeps an action stack so that players have the chance to react to cards played by others using a \texttt{Nope} card. A \texttt{Nope} card cancels the most recent effect, but can itself be cancelled by another \texttt{Nope} card. The turn order and action stack are implemented by extending the base-classes of the framework.  Figure~\ref{fig:ek} shows an example game state.

\subsection{Pandemic~\cite{game:pandemic}}


\textit{Pandemic} is a cooperative board game for $2$ to $4$ players. The board represents a world map, with major cities connected by a graph. Four diseases break out and the objective of the players is to cure them all. Diseases keep spreading after each player's turn, sometimes leading to outbreaks. Each player is assigned a unique role with special abilities and is given cards that can be used for travelling between cities, building research stations or curing diseases. Additionally, they have access to special event cards, which can be played anytime (also out-of-turn). All players lose if they run out of cards in the draw deck, if too many outbreaks occur or if the diseases spread too much. \textit{Pandemic} features partial observability with face-down decks of cards and asymmetric player roles. It employs a reaction system to handle event cards and is the only game currently using the graph-based rule system. 

\begin{figure*}[!ht]
    \centering
    \includegraphics[width=0.8\textwidth]{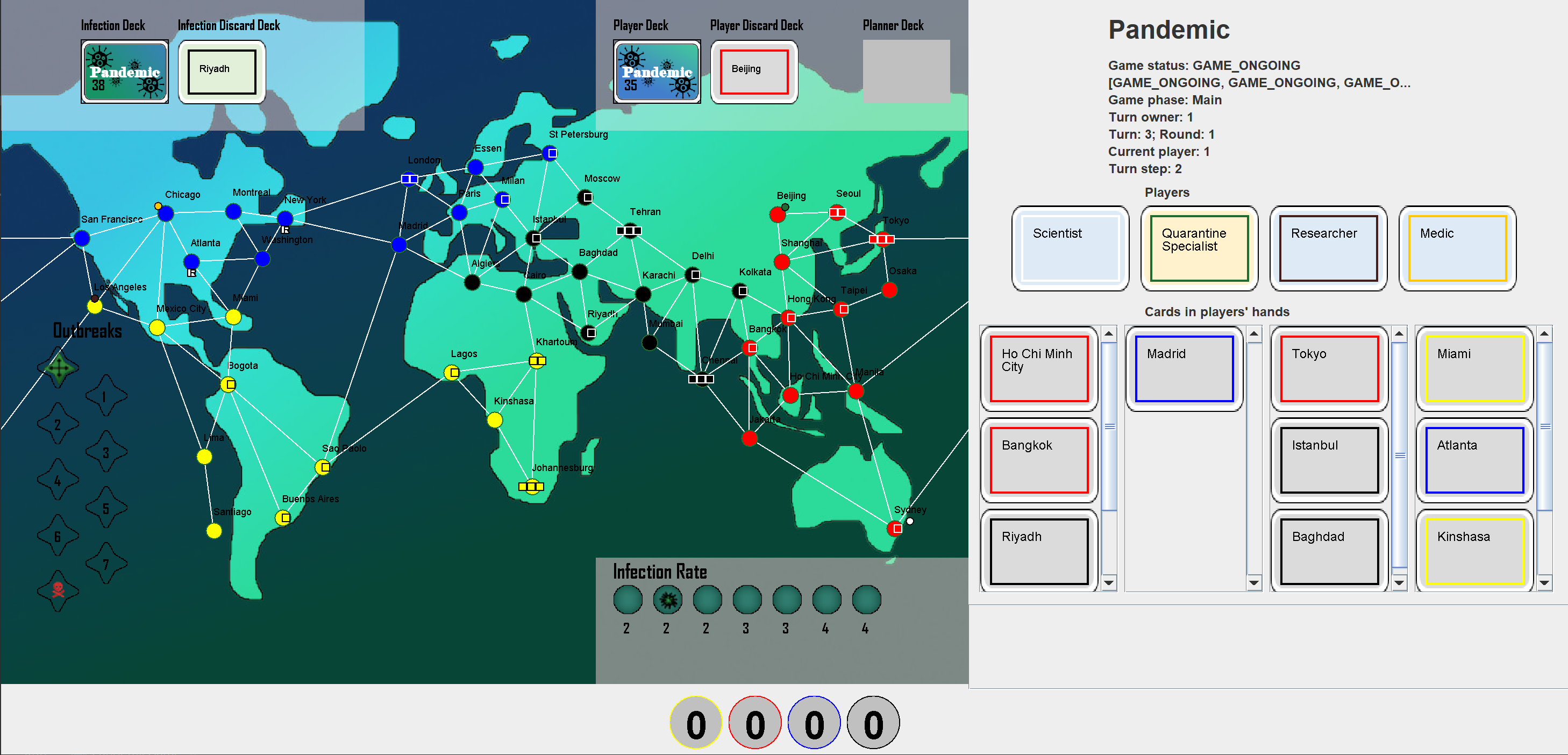}
    \caption{Screenshot of Pandemic in TAG.}
    \label{fig:pan}
\end{figure*}

In each turn, the player can play up to $4$ consecutive actions, with a changing action space (e.g. moving to a new city may result in new actions). To handle event cards and actions that require consent or reactions from other players, a reaction system is used, so that a player can be asked to return an action whenever needed. Pandemic also features partial observability, as the draw deck and the infection deck are not visible.  Figure~\ref{fig:pan} shows an example game state.

\section{Game Analysis}

All games in the framework can be analysed to illustrate the challenge they provide for AI players. Here, we present measurements currently implemented and taken from the existing games described in Section~\ref{sec:games}. We measure the following \textit{averages}, observed in our experiments:

\begin{itemize}
    \item[] $\mu_1$ \textbf{Action space size}: the number of actions available for a player on their turn.
    \item[] $\mu_2$ \textbf{Branching factor}: the number of distinct game states reached through player actions from a given state.
    \item[] $\mu_3$ \textbf{State size}: the number of components in a state.
    \item[] $\mu_4$ \textbf{Amount of hidden information}: the percentage of components hidden from players on their turn.
    \item[] $\mu_5$ \textbf{Game speed}: the execution speed of 4 key functions (in number of calls per second): setup, next, available action computation and state copy.
    \item[] $\mu_6$ \textbf{Game length}: measured as the number of decisions taken by AI players, the total number of game loop iterations (or ticks), the number of rounds in the game and the number of actions per turn (APT) for a player.
    \item[] $\mu_7$ \textbf{Reward sparsity}: granularity of the heuristic functions provided by the game, measured by minimum, maximum and standard deviation of rewards observed by the players.
\end{itemize}

\begin{table*}[!ht]
\centering
\caption{Analysis of games, played 1000 times for each possible number of players on each game, using random agents: action space size, branching factor, state size, hidden information and game speed.}\label{tab:game-measures1}

\begin{tabular}{c|c|c|c|c|c|c|c|c|}
\cline{2-9}
\multirow{2}{*}{} & \multirow{2}{*}{$\mu_1$} & \multirow{2}{*}{$\mu_2$} & \multirow{2}{*}{$\mu_3$} & \multirow{2}{*}{$\mu_4$} & \multicolumn{4}{c|}{$\mu_5$} \\ \cline{6-9}
 &  &  &  &  & Setup & Next & Actions & Copy \\ \hline\hline
\multicolumn{1}{|c|}{Tic-Tac-Toe} & 5.69 & 6.79 & 1 & 0\% & $10^5$ & $10^6$ & $10^5$ & $10^6$ \\ \hline
\multicolumn{1}{|c|}{Love Letter} & 4.74 & 10.78 & 24.00 & 62.96\% & $10^5$ & $10^6$ & $10^5$ & $10^6$ \\ \hline
\multicolumn{1}{|c|}{Uno} & 1.88 & 4.32 & 116.00 & 92.3\% & $10^4$ & $10^6$ & $10^4$ & $10^6$ \\ \hline
\multicolumn{1}{|c|}{Virus!} & 9.70 & 11.03 & 78.00 & 89.1\% & $10^4$ & $10^6$ & $10^5$ & $10^5$ \\ \hline
\multicolumn{1}{|c|}{Exploding Kittens} & 3.17 & 4.99 & 60.00 & 84.5\% & $10^5$ & $10^6$ & $10^5$ & $10^6$ \\ \hline
\multicolumn{1}{|c|}{Colt Express} & 2.91 & 5.46 & 87.60 & 87.67\% & $10^4$ & $10^6$ & $10^4$ & $10^6$ \\ \hline
\multicolumn{1}{|c|}{Pandemic} & 11.15 & 17.97 & 138.00 & 64.97\% & $10^2$ & $10^5$ & $10^4$ & $10^5$ \\ \hline
\end{tabular}

\end{table*}

\begin{table*}[!ht]
\centering
\caption{Analysis of games, played 1000 times for each possible number of players on each game, using random agents: game length and reward sparsity}\label{tab:game-measures2}
\begin{tabular}{c|c|c|c|c|rll|}
\cline{2-8}
\multirow{2}{*}{} & \multicolumn{4}{c|}{$\mu_6$} & \multicolumn{3}{c|}{\multirow{2}{*}{$\mu_7$}} \\ \cline{2-5}
& \#decisions & \#ticks & \#rounds & \#APT & \multicolumn{3}{c|}{}  \\ \hline\hline
\multicolumn{1}{|c|}{Tic-Tac-Toe} & 7.24 & 7.61 & 3.17 & 1 & [\hphantom{-}0.000,& 0.103]& sd=0.02 \\ \hline
\multicolumn{1}{|c|}{Love Letter} & 53.22 & 109.48 & 6.89 & 1.96 & [\hphantom{-}0.000,& 0.970]& sd=0.20 \\ \hline
\multicolumn{1}{|c|}{Uno} & 193.51 & 540.78 & 6.07 & 1 & [-0.150,& 0.270]& sd=0.05  \\ \hline
\multicolumn{1}{|c|}{Virus!} & 317.09 & 319.56 & 75.75 & 1 & [\hphantom{-}0.000,& 0.800]& sd=0.17 \\ \hline
\multicolumn{1}{|c|}{Exploding Kittens} & 51.86 & 73.23 & 8.98 & 1.07 & [-0.500,& 0.750] & sd=0.11 \\ \hline
\multicolumn{1}{|c|}{Colt Express} & 91.71 & 176.13 & 5.00 & 1.00 & [-0.500,& 0.500] &sd=0.14 \\ \hline
\multicolumn{1}{|c|}{Pandemic} & 108.62 & 173.94 & 8.25 & 5.58 & [-1.000,& 0.420] &sd=0.12 \\ \hline
\end{tabular}
\end{table*}

Results are presented in Tables~\ref{tab:game-measures1} and~\ref{tab:game-measures2} (\textit{Virus!} games were limited to 100 rounds, as random play can lead to infinite games). The first thing to note is that all games are very fast to execute: most games can execute over 1 million calls per second to the (usually) most expensive functions (\texttt{next} and \texttt{copy}). The games vary in length, with only $7.61$ ticks for the simplest game, \textit{Tic-Tac-Toe}, but $540.78$ for \textit{Uno}. We further see variations in the state size, with \textit{Pandemic} showing most complex, while \textit{Uno} includes the most hidden information. \textit{Love Letter} shows its strategic complexity through the higher branching factor, while \textit{Exploding Kittens} boasts one of the largest spread of rewards. More complete information and graphical visualisations can be obtained by running the \texttt{evaluation.GameReport} class included with the framework.

\begin{figure}[!ht]
	\centering
    \includegraphics[width=0.75\columnwidth]{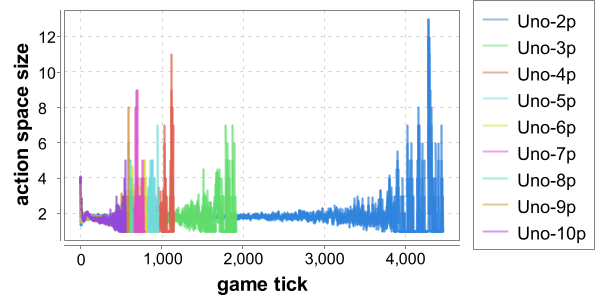}
    \caption{Action space size in \textit{Uno} with all player number versions; 1000 runs per version played by random players.}
    \label{fig:ass-uno}
\end{figure}

Many of the metrics reported do not paint a complete picture if only their average is given: action spaces, for example, vary widely during play for most of the games presented. An example which highlights this is \textit{Uno}, where the action space is dependent on the number of cards in a player's hand, increasing on average as the game goes on (see Figure~\ref{fig:ass-uno}). It is further interesting to note here that games get shorter with more players (as players would earn more points per round if they have more opponents, and thus more cards to total the sum of) - an example of insights which can be obtained through analysis of the games themselves, readily available for any newly implemented games in the framework.

\section{Agents and Algorithms}
\label{sec:algorithm-and-agents}

All implemented players follow a simple interface, only requiring one method to be implemented: \texttt{getAction}. This receives a game state object reduced to the specific player's observation of the current state of the game. How this reduced game state is built is game-dependent, usually randomising unknown information. This method expects an action to be returned out of those available and is called whenever it is the player's turn and they have more than 1 action available (i.e. the player actually has a decision to make). If no decision is required, the agent can choose to still receive and process the information on the game state (in the \texttt{registerUpdatedObservation} function) but an action is not requested. They may also choose to implement the \texttt{initializePlayer} and \texttt{finalizePlayer} functions which are called at the beginning and end of the game, respectively. 

Each player has a player ID assigned by the game engine, and they receive the forward model of the game currently being played. The FM can then be used to advance game states given actions, compute actions available, or reset a game to its initial state. The rest of this section defines the sample players implemented in the framework. These agents use the game's score to evaluate game states (as implemented on the game side and accessible through the \texttt{observation.getScore(this.getPlayerID())} method call), but their heuristic functions may be swapped with a different object implementing the \texttt{IStateHeuristic} interface.

The rest of this section describes the different agents implemented for TAG. Performance of the AI agents described here can be found at~\cite{TAGEXAG20}

\subsection{Human Players}

Two types of human interaction are available, both of which interrupt the game loop to wait for human actions on their turn. \textbf{Console} allows human play using the console. It outputs the game state and available actions in the console and the player inputs the index of the action they choose to play. \textbf{GUI} allows human play with a Graphical User Interface, which is game-specific. It uses an \texttt{ActionController} object to register player action requests, which are then executed in the game.

\subsection{Random}

The simplest automatic player chooses random actions out of those available on its turn. Its implementation is as follows:

\begin{lstlisting}
public class RandomPlayer extends AbstractPlayer {
    private final Random rnd;

    public RandomPlayer()
    {
        this(new Random());
    }

    @Override
    public AbstractAction getAction(AbstractGameState observation) {
        int randomAction = rnd.nextInt(observation.getActions().size());
        List<AbstractAction> actions = observation.getActions();
        return actions.get(randomAction);
    }
}
\end{lstlisting}

\subsection{One Step Look Ahead (OSLA)}

A greedy exhaustive search algorithm, it evaluates all actions from a given game state and picks that which leads to the highest valued state.

\subsection{Rolling Horizon Evolutionary Algorithm (RHEA)}

RHEA~\cite{perez2013rheaptsp} evolves a sequence of $L=10$ actions over several generations, choosing the first action of the best sequence found to play in the game. The algorithm is randomly initialised with a sequence of actions. At each generation it creates a mutation of the current best solution, keeping the best solution of the two. This process repeats until the given budget is exhausted.

Given the variable action spaces and that actions available are highly dependent on the current game state, the mutation operator chooses a gene in the individual (i.e. position in the action sequence) and changes all actions from that point until the end of the individual to new random valid actions. The game's forward model is therefore used in both mutation (to advance the game state given the last action, in order to find the available actions for the given position in the sequence) and evaluation (game states reached through the sequence of actions are evaluated using the game's heuristic, added up for a discounted total with discount factor $\gamma=0.9$, and this total becomes the fitness of the individual). It is important to note that RHEA evolves only its own actions and opponents are given a random model (with intermediate states after opponent actions ignored in fitness evaluations).

\subsection{Monte Carlo Tree Search (MCTS)}

MCTS \cite{browne2012survey} incrementally builds an asymmetric game tree balanced towards to most promising parts of the game state space. It uses multiple iterations of four steps: first, it navigates through the tree, using a \textit{tree policy}, until reaching a node which is not yet fully expanded; next, it adds a new random child of this node to the tree; it then performs a Monte Carlo rollout from the new child (randomly sampling actions until the end of the game or a predetermined depth $L=10$); the state reached at the end of the rollout is evaluated with a heuristic function, and this score is backed up through all the nodes visited during the iteration. The process is repeated until the budget is exhausted, and the most visited child of the root is chosen as the action to play.

The version implemented in the framework is closed-loop: it stores game states in each of the nodes. Further, the rollout step was removed after initial experiments showing an increased performance without it; therefore, the forward model of the game is only used when expanding a leaf node. The resulting node is immediately evaluated using the heuristic and its value is backed up through the tree.


\section{Running the framework}\label{sec:run}

This section describes how to run the games/AI in the framework and the different modes available.

\subsection{Running a game}

The main method in the \texttt{core.Game} class allows for running any game registered in the framework (i.e. those available in the \texttt{games.GameType} enum).

\paragraph{1. Choose on/off visuals:}

Set the ActionController ac variable to null if running without visuals, or leave as-is to enable visuals (if the game has a GUI implemented).

\paragraph{2. Setting the random seed for the game}

Set the seed variable to the specific seed you wish to run the game with (or leave to \texttt{System.currentTimeMillis()} for a new random value):\\
\texttt{long seed = System.currentTimeMillis();}

\paragraph{3. Choose players for the game}

Add to the players array instances of the agents you wish to run the game with. \texttt{HumanConsolePlayer} can be used to play as a human interacting with the game via the console. \texttt{HumanGUIPlayer} can be used to play as a human interacting with the game via the GUI (if implemented).

\begin{lstlisting}
        ArrayList<AbstractPlayer> players = new ArrayList<>();
        players.add(new RandomPlayer(new Random()));
        players.add(new RandomPlayer(new Random()));
        players.add(new RandomPlayer(new Random()));
        players.add(new OSLA());
\end{lstlisting}
        
\paragraph{4. Run!}

Choose final parameters for the method call:

\begin{lstlisting}
private static Game runOne(GameType gameToPlay, List<AbstractPlayer> players, long seed, ActionController ac, boolean randomizeParameters);
\end{lstlisting}

\begin{itemize}
    \item \textbf{gameToPlay}: the type of the game to play
    \item \textbf{players}: previously defined players array
    \item \textbf{seed}: previously defined seed variable
    \item \textbf{ac}: previously defined ac variable
    \item \textbf{randomizeParameters}: true if game parameters should be randomized (if implemented for the game), false otherwise.
\end{itemize}

Adjust settings for the following variables in class \texttt{core.CoreConstants} if needed:

\begin{lstlisting}
    public final static boolean VERBOSE = true;
    public final static boolean PARTIAL_OBSERVABLE = false;
    public final static boolean DISQUALIFY_PLAYER_ON_ILLEGAL_ACTION_PLAYED = false;
\end{lstlisting}

Finally, run the \texttt{core.Game} class.

\subsection{Running multiple games}

To run multiple games, choose one of the following methods, depending on whether you wish to run with a fixed random seed for all games (1, passing desired seed to the method), with a new random seed for each run (1, passing null to the seed parameter in the method) or with fixed random seed for each repetition of each game (2):

\begin{lstlisting}
1. private static void runMany(List<GameType> gamesToPlay, List<AbstractPlayer> players, Long seed, int nRepetitions, ActionController ac, boolean randomizeParameters)
2. private static void runMany(List<GameType> gamesToPlay, List<AbstractPlayer> players, int nRepetitions, long[] seeds, ActionController ac, boolean randomizeParameters)
\end{lstlisting}

Use the chosen function and parameters in the \texttt{core.Game.main} function as before, and run the \texttt{Game} class.

\textbf{Note:} Games within a specific category can be obtained by using the corresponding \texttt{Category} method, e.g. \texttt{GameType.Category.Strategy.getAllGames()} would return a list of all games in the ``Strategy" category.
Games using a specific mechanic can be obtained by using the corresponding \texttt{Mechanic} method, e.g. \texttt{GameType.Mechanic.Cooperative.getAllGames()} would return a list of all cooperative games.

\subsection{Running tournaments}

Tournaments can be run using classes available in the \texttt{evaluation} package. At the time of writing a round-robin tournament is available, which requires the same setup as running single or multiple games in the \texttt{core.Game} class, but will pit all the specified players against all others and run several instances of the given games.

\section{Future Directions}
\label{sec:future-directions}

The presented framework opens up several directions of research and proposes a variety of challenges for AI players, be it search/planning or learning algorithms. Its main focus is to promote research into General Game AI that is able to play many tabletop games at, or surpassing, human level. Relatedly, the agents should be able to handle both \textbf{competitive} (most common testbeds in literature), \textbf{cooperative} and even \textbf{mixed} games. For instance, a future planned development is the inclusion of the game \textit{Betrayal at House on the Hill}~\cite{game:betrayal}, in which the players start off playing cooperatively to later split into teams mid-way through the game, from which point on they are competing instead with newly given team win conditions and rules. Most tabletop games include some degree of \textbf{hidden information} (e.g. face-down decks of cards) and many more players compared to traditional video-game AI testbeds, introducing higher levels of uncertainty. However, such games often make use of similar mechanics, even if in different forms: thus \textbf{knowledge transfer} would be a fruitful area to explore, so that AI players can pick up new game rules more easily based on previous experiences, similar to how humans approach the problem. Some tabletop games further feature \textbf{changing rules} (e.g. \textit{Fluxx}~\cite{game:fluxx}) which would require highly adaptive AI players, able to handle changes in the game engine itself, not only the game state. Many others rely on large amounts of content and components, for which the process of creating new content or modifying the current one for balance, improved synergies etc. could be improved with the help of Procedural Content Generation methods (e.g. cards for the game \textit{Magic the Gathering}~\cite{game:magic} were previously generated in a mixed-initiative method by \cite{summerville2016mystical}).

Specific types of games can also be targeted by research, an option highlighted by TAG's categorisation and labelling of games and their mechanics. Thus AI players could learn to specialise in games using certain mechanics or in areas not yet explored, such as \textbf{Role-Playing} or \textbf{Campaign} games (i.e. games played over several linked and progressive sessions). These games often feature \textbf{asymmetric player roles}, with a special one highlighted (the dungeon master) whose aim is to control the enemies in the game in order to not necessarily win, but give the rest of the players the best experience possible and the right level of challenge. Strategy AI research could see important applications in this domain, as many tabletop games include elements of strategic planning. Role-playing games focused more on the story created by players (e.g. \textit{Dungeons and Dragons}) rather than combat mechanics (e.g. \textit{Gloomhaven}) would also be a very engaging and difficult to approach topic for AI players, where Natural Language Processing research could take an important role.

The framework enables research into \textbf{parameter optimisation}: all parameter classes for games, AI players or heuristics can implement the \texttt{ITunableParameters} interface; parameters can then be automatically randomised, or more intelligently tuned by any optimisation algorithm. This allows for quick and easy exploration of various instances of a problem, a potential increase in AI player performance, or adaptation of AI player behaviour to user preference.

We have mentioned previously that the games implemented offer reduced observations of the game state to the AI players, based on what they can currently observe. These hidden information states (usually) do not keep a history of what was previously revealed to a player. Instead, the AI players should learn to memorise relevant information and build \textbf{belief systems}, as humans would in a real-world context - a very interesting direction of research encouraged by TAG.

Lastly, the framework includes the possibility for games to define their states in terms of either \textbf{vector observations} (\texttt{IVectorObservation}), which enables learning algorithms to be easily integrated with the framework; or \textbf{feature-based observations} (\texttt{IFeatureRepresentation}), which allows for more complex algorithms which can perform a search in the feature space of a game, rather than the usual game state space approached.

\section*{Acknowledgements}
This work was partly funded by the EPSRC CDT in Intelligent Games and Game Intelligence (IGGI)  EP/L015846/1 and EPSRC research grant EP/T008962/1.

\bibliographystyle{unsrt}  
\bibliography{references}

\clearpage

\begin{center}
    \textbf{Appendices}
\end{center}

\appendix

\section{A Game Example: Love Letter} \label{app:loveletter}

This section shows some important code snippets from the implementation of \textit{Love Letter} in TAG.

\subsection[Game State]{Game State: LoveLetterGameState.java} \label{app:loveletter:gs}

\textit{Love Letter} adds one game phase on top of default phases:

\begin{lstlisting}
public enum LoveLetterGamePhase implements IGamePhase {
    Draw
}
\end{lstlisting}

The game state class defines several different components:

\begin{lstlisting}
// List of cards in player hands
List<PartialObservableDeck<LoveLetterCard>> playerHandCards;

// Discarded cards
List<Deck<LoveLetterCard>> playerDiscardCards;

// Cards in draw pile
PartialObservableDeck<LoveLetterCard> drawPile;

// Cards in the reserve
PartialObservableDeck<LoveLetterCard> reserveCards;

// If true: player cannot be effected by any card effects
boolean[] effectProtection;

// Affection tokens per player
int[] affectionTokens;
\end{lstlisting}

and a constructor that initializes the parameters, turn order and game phase:

\begin{lstlisting}
public LoveLetterGameState(AbstractParameters gameParameters, int nPlayers) {
    super(gameParameters, new LoveLetterTurnOrder(nPlayers));
    gamePhase = Draw;
}
\end{lstlisting}

As indicated in Section~\ref{sec:imp:gs}, the following methods are implemented to retrieve all components:

\begin{lstlisting}
@Override
protected List<Component> _getAllComponents() {
    List<Component> components = new ArrayList<>();
    components.addAll(playerHandCards);
    components.addAll(playerDiscardCards);
    components.add(drawPile);
    components.add(reserveCards);
    return components;
}
\end{lstlisting}

copy a game state for a particular player (defined by a \texttt{playerId} variable):

\begin{lstlisting}
@Override
protected AbstractGameState _copy(int playerId) {
    LoveLetterGameState llgs = new LoveLetterGameState(gameParameters.copy(), getNPlayers());
    llgs.drawPile = drawPile.copy();
    llgs.reserveCards = reserveCards.copy();
    llgs.playerHandCards = new ArrayList<>();
    llgs.playerDiscardCards = new ArrayList<>();
    for (int i = 0; i < getNPlayers(); i++) {
        llgs.playerHandCards.add(playerHandCards.get(i).copy());
        llgs.playerDiscardCards.add(playerDiscardCards.get(i).copy());
    }
    llgs.effectProtection = effectProtection.clone();
    llgs.affectionTokens = affectionTokens.clone();

    if (PARTIAL_OBSERVABLE && playerId != -1) {
        // Draw pile, some reserve cards and other player's hand is possibly hidden. Mix all together and draw randoms
        for (int i = 0; i < getNPlayers(); i++) {
            if (i != playerId && llgs.playerHandCards.get(i).getDeckVisibility()[playerId]) {
                // Hide!
                llgs.drawPile.add(llgs.playerHandCards.get(i));
                llgs.playerHandCards.get(i).clear();
            }
        }
        for (int i = 0; i < llgs.reserveCards.getSize(); i++) {
            if (!llgs.reserveCards.isComponentVisible(i, playerId)) {
                // Hide!
                llgs.drawPile.add(llgs.reserveCards.get(i));
            }
        }
        Random r = new Random(llgs.getGameParameters().getRandomSeed());
        llgs.drawPile.shuffle(r);
        for (int i = 0; i < getNPlayers(); i++) {
            if (i != playerId && llgs.playerHandCards.get(i).getDeckVisibility()[playerId]) {
                // New random cards
                for (int j = 0; j < playerHandCards.get(i).getSize(); j++) {
                    llgs.playerHandCards.get(i).add(llgs.drawPile.draw());
                }
            }
        }
        for (int i = 0; i < llgs.reserveCards.getSize(); i++) {
            if (!llgs.reserveCards.isComponentVisible(i, playerId)) {
                // New random card
                llgs.reserveCards.setComponent(i, llgs.drawPile.draw());
            }
        }
    }
    return llgs;
}
\end{lstlisting}

The function \texttt{reset()} sets the data structures of the game state to a functional but empty state:

\begin{lstlisting}
@Override
protected void _reset() {
    gamePhase = Draw;
    playerHandCards = new ArrayList<>();
    playerDiscardCards = new ArrayList<>();
    drawPile = null;
    reserveCards = null;
    effectProtection = new boolean[getNPlayers()];
}
\end{lstlisting}

and the \texttt{getScore()} function uses a helper method to determine the heuristic value of this state:

\begin{lstlisting}
@Override
protected double _getScore(int playerId) {
    return new LoveLetterHeuristic().evaluateState(this, playerId);
}
\end{lstlisting}

This heuristic method is defined as follows in \texttt{games.loveletter.LoveLetterHeuristic.java}:

\begin{lstlisting}
public double evaluateState(AbstractGameState gs, int playerId) {
    LoveLetterGameState llgs = (LoveLetterGameState) gs;
    LoveLetterParameters llp = (LoveLetterParameters) gs.getGameParameters();
    Utils.GameResult playerResult = gs.getPlayerResults()[playerId];

    if (playerResult == Utils.GameResult.LOSE)
        return -1;
    if (playerResult == Utils.GameResult.WIN)
        return 1;

    double cardValues = 0;

    Random r = new Random(llgs.getGameParameters().getRandomSeed());
    for (LoveLetterCard card: llgs.getPlayerHandCards().get(playerId).getComponents()) {
        if (card.cardType == LoveLetterCard.CardType.Countess) {
            if (r.nextDouble() > COUNTESS_PLAY_THRESHOLD) {
                cardValues += LoveLetterCard.CardType.Countess.getValue();
            }
        } else {
            cardValues += card.cardType.getValue();
        }
    }

    double maxCardValue = 1+llgs.getPlayerHandCards().get(playerId).getSize() * LoveLetterCard.CardType.getMaxCardValue();
    double nRequiredTokens = (llgs.getNPlayers()-1 < llp.nTokensWin.length ? llp.nTokensWin[llgs.getNPlayers()-1] :
            llp.nTokensWin[llp.nTokensWin.length-1]);
    if (nRequiredTokens < llgs.affectionTokens[playerId]) nRequiredTokens = llgs.affectionTokens[playerId];

    return FACTOR_CARDS * (cardValues/maxCardValue) + FACTOR_AFFECTION * (llgs.affectionTokens[playerId]/nRequiredTokens);
}
\end{lstlisting}

\subsection[Forward Model]{Forward Model: LoveLetterForwardModel.java} \label{app:loveletter:fm}

The first game state is initialised in the \texttt{\_setup()} method:

\begin{lstlisting}
@Override
protected void _setup(AbstractGameState firstState) {
    LoveLetterGameState llgs = (LoveLetterGameState)firstState;

    // Set up all variables
    llgs.drawPile = new PartialObservableDeck<>("drawPile", llgs.getNPlayers());
    llgs.reserveCards = new PartialObservableDeck<>("reserveCards", llgs.getNPlayers());
    llgs.affectionTokens = new int[llgs.getNPlayers()];
    llgs.playerHandCards = new ArrayList<>(llgs.getNPlayers());
    llgs.playerDiscardCards = new ArrayList<>(llgs.getNPlayers());

    // Set up first round
    setupRound(llgs, null);
}
\end{lstlisting}

This setup counts on the auxiliary method setupRound(). This method is separate as it is called whenever a new round starts in the game, and it sets up a round for the game, including the draw pile, reserve pile and starting player hands:

\begin{lstlisting}
private void setupRound(LoveLetterGameState llgs, HashSet<Integer> previousWinners) {
    LoveLetterParameters llp = (LoveLetterParameters) llgs.getGameParameters();

    // No protection this round
    llgs.effectProtection = new boolean[llgs.getNPlayers()];

    // Reset player status
    for (int i = 0; i < llgs.getNPlayers(); i++) {
        llgs.setPlayerResult(Utils.GameResult.GAME_ONGOING, i);
    }

    // Add all cards to the draw pile
    llgs.drawPile.clear();
    for (HashMap.Entry<LoveLetterCard.CardType, Integer> entry : llp.cardCounts.entrySet()) {
        for (int i = 0; i < entry.getValue(); i++) {
            LoveLetterCard card = new LoveLetterCard(entry.getKey());
            llgs.drawPile.add(card);
        }
    }

    // Put one card to the side, such that player's won't know all cards in the game
    Random r = new Random(llgs.getGameParameters().getRandomSeed() + llgs.getTurnOrder().getRoundCounter());
    llgs.drawPile.shuffle(r);
    llgs.reserveCards.clear();
    llgs.reserveCards.add(llgs.drawPile.draw());

    // In min-player game, N more cards are on the side, but visible to all players at all times
    if (llgs.getNPlayers() == GameType.LoveLetter.getMinPlayers()) {
        boolean[] fullVisibility = new boolean[llgs.getNPlayers()];
        Arrays.fill(fullVisibility, true);
        for (int i = 0; i < llp.nCardsVisibleReserve; i++) {
            llgs.reserveCards.add(llgs.drawPile.draw(), fullVisibility);
        }
    }

    // Set up player hands and discards
    llgs.playerHandCards.clear();
    llgs.playerDiscardCards.clear();
    for (int i = 0; i < llgs.getNPlayers(); i++) {
        boolean[] visible = new boolean[llgs.getNPlayers()];
        if (PARTIAL_OBSERVABLE) {
            visible[i] = true;
        } else {
            Arrays.fill(visible, true);
        }

        // add random cards to the player's hand
        PartialObservableDeck<LoveLetterCard> playerCards = new PartialObservableDeck<>("playerHand" + i, i, visible);
        for (int j = 0; j < llp.nCardsPerPlayer; j++) {
            playerCards.add(llgs.drawPile.draw());
        }
        llgs.playerHandCards.add(playerCards);

        // create a player's discard pile, which is visible to all players
        Deck<LoveLetterCard> discardCards = new Deck<>("discardPlayer" + i, i);
        llgs.playerDiscardCards.add(discardCards);
    }

    // Game starts with drawing cards
    llgs.setGamePhase(Draw);

    if (previousWinners != null) {
        // Random winner starts next round
        int nextPlayer = r.nextInt(previousWinners.size());
        int n = -1;
        for (int i: previousWinners) {
            n++;
            if (n == nextPlayer) {
                llgs.getTurnOrder().setTurnOwner(i);
            }
        }
    }

    // Update components in the game state
    llgs.updateComponents();
}
\end{lstlisting}

The method that rolls the state forward with a given action, \texttt{next()}, is implemented as follows:

\begin{lstlisting}
 @Override
protected void _next(AbstractGameState gameState, AbstractAction action) {
    // each turn begins with the player drawing a card after which one card will be played
    // switch the phase after each executed action
    LoveLetterGameState llgs = (LoveLetterGameState) gameState;
    action.execute(gameState);

    IGamePhase gamePhase = llgs.getGamePhase();
    if (gamePhase == Draw) {
        llgs.setGamePhase(AbstractGameState.DefaultGamePhase.Main);
    } else if (gamePhase == AbstractGameState.DefaultGamePhase.Main) {
        llgs.setGamePhase(Draw);
        llgs.getTurnOrder().endPlayerTurn(gameState);
        checkEndOfRound(llgs);
    } else {
        throw new IllegalArgumentException("The game phase " + llgs.getGamePhase() +
                " is not know by LoveLetterForwardModel");
    }
}
\end{lstlisting}

Another important method in \textit{Love Letter}'s forward model is the one in charge of computing the available actions for a player. The snippet for that method is shown below:

\begin{lstlisting}
@Override
protected List<AbstractAction> _computeAvailableActions(AbstractGameState gameState) {
    LoveLetterGameState llgs = (LoveLetterGameState)gameState;
    ArrayList<AbstractAction> actions;
    int player = gameState.getTurnOrder().getCurrentPlayer(gameState);
    if (gameState.getGamePhase().equals(AbstractGameState.DefaultGamePhase.Main)) {
        actions = playerActions(llgs, player);
    } else if (gameState.getGamePhase().equals(LoveLetterGameState.LoveLetterGamePhase.Draw)) {
        // In draw phase, the players can only draw cards.
        actions = new ArrayList<>();
        actions.add(new DrawCard(llgs.drawPile.getComponentID(), llgs.playerHandCards.get(player).getComponentID(), 0));
    } else {
        throw new IllegalArgumentException(gameState.getGamePhase().toString() + " is unknown to LoveLetterGameState");
    }

    return actions;
}
\end{lstlisting}

In \textit{Love Letter}, the actions available depend on the card the player holds. Here's an extract of the method \texttt{playerActions()} that \texttt{\_computeAvailableActions()} uses:

\begin{lstlisting}
switch (playerDeck.getComponents().get(card).cardType) {
        case Priest:
            for (int targetPlayer = 0; targetPlayer < llgs.getNPlayers(); targetPlayer++) {
                if (targetPlayer == playerID || llgs.getPlayerResults()[targetPlayer] == Utils.GameResult.LOSE)
                    continue;
                actions.add(new PriestAction(playerDeck.getComponentID(),
                        playerDiscardPile.getComponentID(), card, targetPlayer));
            }
            break;

        case Guard:
            for (int targetPlayer = 0; targetPlayer < llgs.getNPlayers(); targetPlayer++) {
                if (targetPlayer == playerID || llgs.getPlayerResults()[targetPlayer] == Utils.GameResult.LOSE)
                    continue;
                for (LoveLetterCard.CardType type : LoveLetterCard.CardType.values())
                    actions.add(new GuardAction(playerDeck.getComponentID(),
                            playerDiscardPile.getComponentID(), card, targetPlayer, type));
            }
            break;

        case Baron:
            for (int targetPlayer = 0; targetPlayer < llgs.getNPlayers(); targetPlayer++) {
                if (targetPlayer == playerID || llgs.getPlayerResults()[targetPlayer] == Utils.GameResult.LOSE)
                    continue;
                actions.add(new BaronAction(playerDeck.getComponentID(),
                        playerDiscardPile.getComponentID(), card, targetPlayer));
            }
            break;

        case Handmaid:
            actions.add(new HandmaidAction(playerDeck.getComponentID(),
                    playerDiscardPile.getComponentID(), card));
            break;

        case Prince:
            for (int targetPlayer = 0; targetPlayer < llgs.getNPlayers(); targetPlayer++) {
                if (targetPlayer == playerID || llgs.getPlayerResults()[targetPlayer] == Utils.GameResult.LOSE)
                    continue;
                actions.add(new PrinceAction(playerDeck.getComponentID(),
                        playerDiscardPile.getComponentID(), card, targetPlayer));
            }
            break;

        case King:
            for (int targetPlayer = 0; targetPlayer < llgs.getNPlayers(); targetPlayer++) {
                if (targetPlayer == playerID || llgs.getPlayerResults()[targetPlayer] == Utils.GameResult.LOSE)
                    continue;
                actions.add(new KingAction(playerDeck.getComponentID(),
                        playerDiscardPile.getComponentID(), card, targetPlayer));
            }
            break;

        case Countess:
            actions.add(new CountessAction(playerDeck.getComponentID(),
                    playerDiscardPile.getComponentID(), card));
            break;

        case Princess:
            actions.add(new PrincessAction(playerDeck.getComponentID(),
                    playerDiscardPile.getComponentID(), card));
            break;
\end{lstlisting}

\clearpage
\subsection{Actions in Love Letter} \label{app:loveletter:actions}

Here are some actions that are exclusive to the game \textit{Love Letter}. Actions can extend other actions provided within TAG, enhancing their functionality. For instance, the following DrawCard action modifies the execution of a regular drawing card move to set certain values in the game state. 

\begin{lstlisting}
public class DrawCard extends core.actions.DrawCard implements IPrintable {

    public DrawCard(int deckFrom, int deckTo, int fromIndex) {
        super(deckFrom, deckTo, fromIndex);
    }

    @Override
    public boolean execute(AbstractGameState gs) {
        ((LoveLetterGameState)gs).setProtection(gs.getTurnOrder().getCurrentPlayer(gs), false);
        return super.execute(gs);
    }

    @Override
    public String getString(AbstractGameState gameState) {
        return "Draw a card and remove protection status.";
    }

    @Override
    public String toString() {
        return "Draw a card and remove protection status.";
    }

    @Override
    public void printToConsole(AbstractGameState gameState) {
        System.out.println(toString());
    }

    @Override
    public AbstractAction copy() {
        return new DrawCard(deckFrom, deckTo, fromIndex);
    }
}
\end{lstlisting}

This \texttt{DrawCard} action is further extended to provide particular effects for each action in the game. For instance, in \textit{Love Letter}, the \texttt{execute()} function for playing a Baron card is implemented as follows:

\begin{lstlisting}
@Override
public boolean execute(AbstractGameState gs) {
    LoveLetterGameState llgs = (LoveLetterGameState)gs;
    int playerID = gs.getTurnOrder().getCurrentPlayer(gs);
    Deck<LoveLetterCard> playerDeck = llgs.getPlayerHandCards().get(playerID);
    Deck<LoveLetterCard> opponentDeck = llgs.getPlayerHandCards().get(opponentID);

    // compares the value of the player's hand card with another player's hand card
    // the player with the lesser valued card will be removed from the game
    if (llgs.isNotProtected(opponentID) && gs.getPlayerResults()[playerID] != Utils.GameResult.LOSE){
        LoveLetterCard opponentCard = opponentDeck.peek();
        LoveLetterCard playerCard = playerDeck.peek();
        if (opponentCard != null && playerCard != null) {
            if (opponentCard.cardType.getValue() < playerCard.cardType.getValue())
                llgs.killPlayer(opponentID);
            else if (playerCard.cardType.getValue() < opponentCard.cardType.getValue())
                llgs.killPlayer(playerID);
        } else {
            throw new IllegalArgumentException("player with ID " + opponentID + " was targeted using a Baron card" +
                    " but one of the players has now cards left.");
        }
    }

    return super.execute(gs);
}
\end{lstlisting}

\clearpage
Other actions are simpler, such as playing a Guard card:

\begin{lstlisting}
@Override
public boolean execute(AbstractGameState gs) {
    LoveLetterGameState llgs = (LoveLetterGameState)gs;
    Deck<LoveLetterCard> opponentDeck = llgs.getPlayerHandCards().get(opponentID);

    // guess the opponent's card and remove the opponent from play if the guess was correct
    if (llgs.isNotProtected(opponentID)){
        LoveLetterCard card = opponentDeck.peek();
        if (card.cardType == this.cardType) {
            llgs.killPlayer(opponentID);
        }
    }
    return super.execute(gs);
}
\end{lstlisting}

or a Priest card:

\begin{lstlisting}
@Override
public boolean execute(AbstractGameState gs) {
    LoveLetterGameState llgs = (LoveLetterGameState)gs;
    int playerID = gs.getTurnOrder().getCurrentPlayer(gs);
    PartialObservableDeck<LoveLetterCard> opponentDeck = llgs.getPlayerHandCards().get(opponentID);

    // Set all cards to be visible by the current player
    if (((LoveLetterGameState) gs).isNotProtected(opponentID)){
        for (int i = 0; i < opponentDeck.getComponents().size(); i++)
            opponentDeck.setVisibilityOfComponent(i, playerID, true);
    }

    return super.execute(gs);
}
\end{lstlisting}
    
\end{document}